\documentclass[conference]{IEEEtran}

\usepackage{cite}
\usepackage{amsmath,amssymb,amsfonts}
\usepackage{algorithmic}
\usepackage{graphicx}
\usepackage{textcomp}
\usepackage{xcolor}
\usepackage{url}
\usepackage{epsfig}
\usepackage{soul}
\usepackage{multirow}
\usepackage{blindtext}
\usepackage{comment}
\usepackage{multicol}
\usepackage{multirow}
\usepackage{pdfpages}
\usepackage{subfigure}
\usepackage{booktabs}
\usepackage{tabularx}
\usepackage{adjustbox}
\usepackage{glossaries}
\usepackage{arydshln}
\graphicspath{ {./Figures/} }
\usepackage{array,multirow,makecell}
\usepackage{bm}
\usepackage{acronym}
\newacro{dnn}[DNN]{deep neural network}
\newacro{cnn}[CNN]{convolutional neural network}
\newacro{ae}[AE]{Autoencoder}
\newacro{gan}[GAN]{Generative Adversarial Network}

\newcommand{\STAB}[1]{\begin{tabular}{@{}c@{}}#1\end{tabular}}

\def\BibTeX{{\rm B\kern-.05em{\sc i\kern-.025em b}\kern-.08em
    T\kern-.1667em\lower.7ex\hbox{E}\kern-.125emX}}

\begin{document}

\title{Evaluation of Pre-Trained CNN Models for Geographic Fake Image Detection}

\author{\IEEEauthorblockN{Sid Ahmed Fezza$^{1}$, Mohammed Yasser Ouis$^{1}$, Bachir Kaddar$^{2}$, Wassim Hamidouche$^{3}$, Abdenour Hadid$^{4}$}
\IEEEauthorblockA{$^{1}$National Higher School of Telecommunications and ICT, Oran, Algeria \\
$^{2}$University of Ibn Khaldoun, Tiaret, Algeria\\
$^{3}$Univ. Rennes, INSA Rennes, CNRS, IETR - UMR 6164, Rennes, France\\
$^{4}$Sorbonne Center for Artificial Intelligence, Sorbonne University Abu Dhabi, Abu Dhabi, UAE\\
sfezza@ensttic.dz}
}

\maketitle

\begin{abstract}
Thanks to the remarkable advances in generative adversarial networks (GANs), it is becoming increasingly easy to generate/manipulate images. The existing works have mainly focused on deepfake in face images and videos. However, we are currently witnessing the emergence of fake satellite images, which can be misleading or even threatening to national security. Consequently, there is an urgent need to develop detection methods capable of distinguishing between real and fake satellite images. To advance the field, in this paper, we explore the suitability of several convolutional neural network (CNN) architectures for fake satellite image detection. Specifically, we benchmark four CNN models by conducting extensive experiments to evaluate their performance and robustness against various image distortions. This work allows the establishment of new baselines and may be useful for the development of CNN-based methods for fake satellite image detection.
\end{abstract}

 
\section{Introduction}
\label{sec:intro}
Deep learning has proven to be effective in solving various problems in the field of computer vision \cite{chai2021deep}. In particular, with the development of new and robust generative deep learning algorithms \cite{goodfellow2014generative}, a powerful new application has emerged known as \textit{deepfake}. Deepfake consists in generating synthetic media, i.e., images, audios or videos, with manipulated content.

With the introduction of more advanced generative neural networks (such as  \acp{ae}~\cite{vincent2008extracting} or \acp{gan}~\cite{goodfellow2014generative}), the availability of high-performance computational resources (e.g. GPU), and the easy access to a large number of social medias and public databases \cite{korshunov2018deepfakes, rossler2019faceforensics++,dolhansky2020deepfake}, deepfake generation has progressed significantly. Synthetic yet very realistic data generation has indeed improved significantly \cite{ajder2019state}, making it very difficult to distinguish between real and synthesized videos, even for a human observer~\cite{korshunov2021subjective}. 

Despite their various positive applications~\cite{marr2019best}, there is a potential harm and malicious use of deepfake, that can negatively affect privacy, democracy and national security \cite{chesney2019deep,vaccari2020deepfakes}. Hence, the topic of deepfake detection has gained more attention by the research community~\cite{tolosana2020deepfakes,mirsky2021creation}. For instance, the Defense Advanced Research Project Agency (DARPA) initiated a novel research direction aimed at focusing academic research efforts to accelerate the development of new and more robust deepfake detection techniques. The National Institute of Standards and Technology (NIST) and Facebook launched the Media Forensics Challenge (MFC2018) and the Deepfake Detection Challenge (DFDC) to promote the development of advanced solutions that detect AI-generated and manipulated videos. As a result, the number of proposed solutions have increased significantly to address this challenging problem of deepfake detection. \cite{lyu2022deepfake}.  

Satellite images are a valuable and common source of data for cartography and geographic information science, which are important domains that give humans the ability to analyze and explore the real geographic world. Deepfake geography~\cite{zhao2021deep}, on the other hand, has emerged as a recent concern which can pose a national security threat~\cite{tucker2019newest}. The aim of deepfake geography consists of producing falsified satellite images by replacing an image at one location with another image, while still maintaining a realistic appearance for the entire scene~\cite{xu2018satellite}. Despite the remarkable advances in deepfakes detection, research focus on fake satellite images detection is very limited. 

Recently, with the widespread growth of the geographic deepfake problem~\cite{zhongming2021growing}, researchers started looking at detecting whether satellites images have been manipulated or not~\cite{chen2021geo}. Among the first attempts to tackle this problem is the introduction of public fake satellite image dataset~\cite{zhao2021deep}. The authors proposed a \ac{gan}-based image-to-image translation approach to generate fake satellite images. By learning the characteristics of an urban area, a fake satellite image is then generated by mapping the learned characteristics onto a different base map. Moreover, the authors proposed a method based on handcrafted features to determine whether the satellite image is authentic or fake. 
Following this work, another interesting method  called Geo-DefakeHop~\cite{chen2021geo} has been proposed. This method is based on parallel subspace learning (PSL), which consists of multiple filter banks used simultaneously. The authors focused on the higher frequency components to learn the differences
between real and fake satellite images.

In this paper, we evaluate the performance of several deep \ac{cnn} architectures, which have shown state-of-the-art accuracy for general image classification tasks, for fake satellite images detection. We conduct a thorough evaluation using the most promising \acp{cnn} with the aim of establishing new baselines and promoting future research in this under-explored area of geographic fake image detection.

The rest of this paper is structured as follows. The relevant related works are presented in Section~\ref{sec:related}. In Section~\ref{sec:Methodology}, the methodology is described. Experimental results and discussions are given in Section \ref{sec:Exp}, and the conclusions are drawn in Section \ref{sec:conclusion}.
\section{Related work}
\label{sec:related}
Images/videos of faces are by far the most targeted in the fake multimedia field \cite{verdoliva2020media,masood2022deepfakes,dolhansky2020deepfake}, and very few works deal with fake satellite images. Early studies identify deepfakes by detecting specific artifacts that occurred on the most common first-generation manipulated videos \cite{agarwal2017swapped, akhtar2019comparative,matern2019exploiting}. For instance, in \cite{matern2019exploiting}, the authors used statistical differences in color components to distinguish synthetic deepfakes from authentic images. However, it is difficult to handcraft all the most suitable and meaningful features.

Unlike previous deepfake generation techniques, \ac{gan}-based approaches are able to produce more realistic deepfake videos with high visual quality \cite{karras2017progressive,miyato2018spectral,bellemare2017cramer,binkowski2018demystifying}, making the deepfake detection task more challenging. Therefore, more robust methods have been developed that focus on detecting subtle features and inconsistencies between frames brought by the \ac{gan} model \cite{rossler2018faceforensics}. For instance, in \cite{mccloskey2018detecting}, the authors proposed a detection method based on color features and a linear Support Vector Machine (SVM) for the final classification, achieving promising results on the NIST MFC2018 dataset \cite{guan2019mfc}. Zhang \textit{et al.} \cite{zhang2017automated} used the bag-of-words method to extract a set of discriminating features that were fed  into various classifiers to discriminate manipulated face images from real ones. A significant drawback of these approaches is that the models have to be re-trained again when simple unseen image perturbation attacks (such as noise, blur, cropping or compression) are presented.

In recent years, it has been shown that deep neural networks tend to perform better than traditional image forensic tools \cite{bayar2016deep,rahmouni2017distinguishing,marra2018detection}. Consequently, most developed methods adopted it to automatically extract discriminating features to detect deepfakes. These CNN-based approaches mainly rely on the presence of visual artifacts or inconsistency of intrinsic features between fake and real images/videos \cite{tariq2018detecting}. These artifacts can be detected by well established and pre-trained CNN models such as VGG16 \cite{simonyan2014very}, ResNet50 \cite{he2016deep}, and so on. For instance, Rossler \textit{et al.} \cite{rossler2019faceforensics++} used XceptionNet model that was trained on the Faceforensics++ dataset to detect manipulated images.  Nguyen \textit{et al.} \cite{nguyen2019capsule} proposed the use of capsule networks that takes features obtained from the VGG-19 network \cite{simonyan2014very} to detect forged images and videos. Marra \textit{et al.} performed in \cite{marra2019incremental} an interesting study in order to classify correctly unseen types of GAN-based fake generated data. The authors proposed a multi-task incremental learning detection method based on the XceptionNet model and achieved promising results. 

Despite the progress made on deepfake detection methods in face images and videos, research on fake satellite image detection is still limited, mainly due to the lack of labeled fake satellite image datasets. In~\cite{zhao2021deep}, the authors draw the public's attention to the potential danger inherent in the progressive evolution of this technology (GeoAI capabilities) and its impact on our perception of the geographic world. For this purpose, they recently proposed the first fake satellite image dataset using a GAN-based image-to-image translation approach. This dataset consists of falsified satellite images from three cities, including Tacoma, Seattle and Beijing. Furthermore, the authors identified 26 salient hand-crafted features that show a significant value difference between authentic and fake satellite images. They classified these features into three classes, which are spatial, histogram and frequency features. These salient features are concatenated and fed into an SVM classifier trained to differentiate fake satellite images from the authentic ones. Chen \textit{et al.}~\cite{chen2021geo} introduced a learning method for the detection of fake satellite images called Geo-DefakeHop. The idea behind their approach is that GANs can generate high-quality realistic images by reproducing well the low-frequency responses of synthesized images. However, it cannot do the same for high-frequency components due to the complexity constraint. Thus, the authors proposed to use multiple filter banks, named parallel subspace learning (PSL), selected a few discriminant features from each filter bank and grouped them for binary classification.
\begin{table}[t!]
\centering
\caption{The characteristics of the four considered CNNs pre-trained models.}
\label{tab::keras}
\begin{adjustbox}{max width=0.5\textwidth}
\begin{tabular}{|r|c|c|c|c|}
\hline
\textbf{Model}                               & \textbf{Parameters} & \textbf{Depth} & \textbf{Top-1 Accuracy} & \textbf{Top-5 Accuracy} \\ \hline
ResNet50\cite{he2016deep}                    & 25,636,712                           & 168                             & 0.749                                    & 0.921                                    \\ \hline
VGG16\cite{simonyan2014very}                 & 138,357,544                          & 23                              & 0.713                                    & 0.901                                    \\ \hline
InceptionV3\cite{szegedy2016rethinking}      & 23,851,784                           & 159                             & 0.779                                    & 0.937                                    \\ \hline
Xception\cite{chollet2017xception}                      & 22,910,480                           & 126                             & 0.790                                    & 0.945                                    \\ \hline
\end{tabular}
\end{adjustbox}
\end{table}
\section{Methodology}
\label{sec:Methodology}
In this section, we start by briefly introducing the considered CNN architectures. Then, we discuss the details of training and tuning these models using the transfer learning technique for the fake satellite image detection task.
\subsection{Deep Convolutional Neural Network Architectures}
\label{sec:CNN}
Taking advantage of the proliferation of large datasets in addition to increasing in computational power, deep neural networks (DNNs) have become the most efficient methods for many computer vision tasks. For instance, in some visual recognition tasks, the DNNs are able to recognize images better than humans. 

Over the past few years, many DNNs models have been proposed and have shown top results on the ImageNet dataset \cite{russakovsky2015imagenet}, which represents the large-scale image classification dataset. Among them, we can cite VGG16 \cite{simonyan2014very}, InceptionV3 \cite{szegedy2016rethinking}, ResNet50 \cite{he2016deep} and Xception  \cite{chollet2017xception}, which we considered in this study. The number of parameters and the depth of each model are summarized in Table \ref{tab::keras}. In addition, top-1 and top-5 accuracy that refers to model's performance on the ImageNet validation dataset are also provided.
\begin{figure}[t!]
    \centering
    \includegraphics[width=0.7\linewidth]{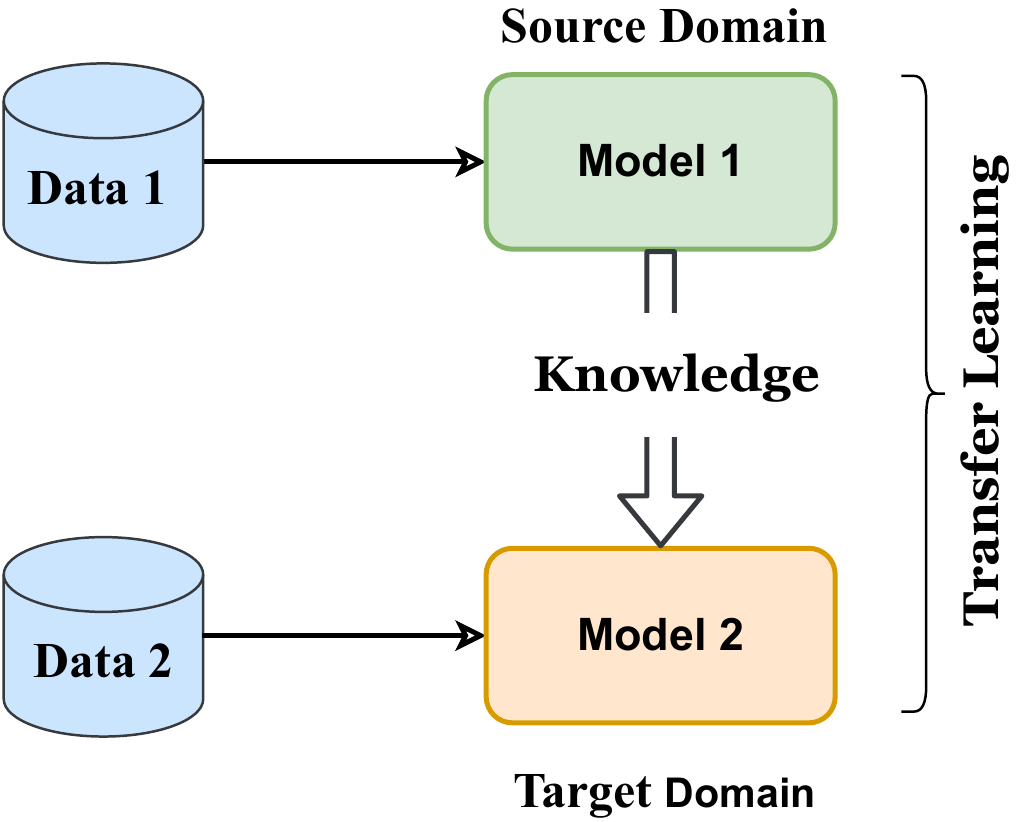}
    \caption{Learning process of transfer learning.}
    \label{fig:tf}
\end{figure}
\begin{figure*}[t!]
\centering
\subfigure{\includegraphics[width=0.28\linewidth]{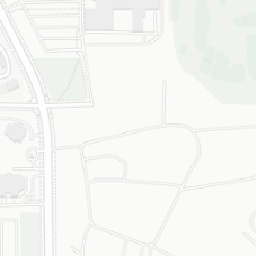}}
\subfigure{\includegraphics[width=0.28\linewidth]{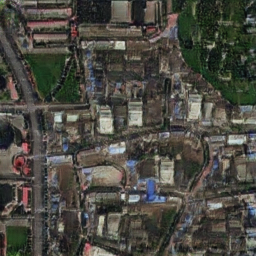}}\hspace{0.0001em}
\subfigure{\includegraphics[width=0.28\linewidth]{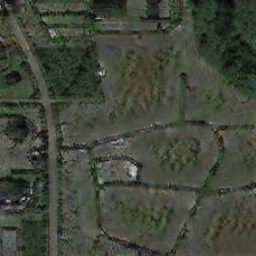}}\\
\addtocounter{subfigure}{-3}
\subfigure[Base map]{\includegraphics[width=0.28\linewidth]{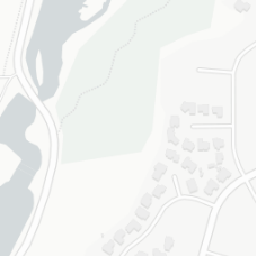}}
\subfigure[Beijing]{\includegraphics[width=0.28\linewidth]{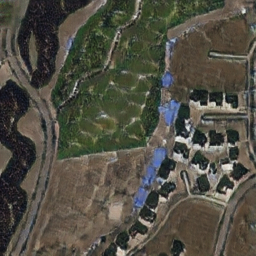}}\hspace{0.0001em}
\subfigure[Seattle]{\includegraphics[width=0.28\linewidth]{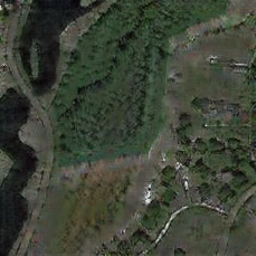}}\hspace{0.0001em}
\caption{Some samples of fake satellite images from the UW dataset \cite{zhao2021deep}. A neighborhood in Tacoma with landscape features from other cities. (a) The CartoDB base map of Tacoma, fake satellite images generated with the visual patterns of (b) Beijing and (c) Seattle.}
 \label{fig:uw}
\end{figure*}

Following the success of the AlexNet \cite{krizhevsky2012imagenet} model, VGG network \cite{simonyan2014very} was the first architecture that relied on small size filters, which improves the CNN performance. This is due to the fact that two $3\times3$ convolutions layers have the same receptive field as a single $5\times5$ convolution, while having fewer parameters and taking less computation. VGGNet has two variants, VGG16 and VGG19, which represent the number of convolutional layers in the model. InceptionV3 \cite{szegedy2016rethinking} is based on some of the original ideas of GoogleNet \cite{szegedy2015going} and VGGNet. In InceptionV3, several techniques for optimizing the network have been exploited, including factorized convolutions, regularization, dimension reduction, and parallelized computations. Deep residual networks \cite{he2016deep}, shortened ResNet, introduced a concept that allowed the development of much deeper networks. Deeper networks can learn more complex functions and representations of the inputs that usually lead to better performance. However, a deep network still faces the challenges of network degradation as well as exploding or vanishing gradients. In ResNet, residual blocks were introduced, in which  the inputs are added back to their outputs with the aim of creating identity mappings. Such identity mappings are performed using the so-called skip or residual connections. Xception \cite{chollet2017xception} takes the principles of Inception to the extreme. The main idea behind Xception is its depthwise separable convolution, i.e., a depthwise convolution followed by a pointwise convolution, which provides better performance than InceptionV3. Xception adjusted the original inception block by mapping the spatial correlations separately for each output channel instead of partitioning input data into multiple compressed chunks and then performing a $1\times1$ depth conversion to capture cross-channel correlation.
\subsection{Transfer Learning for Geographic Fake Image Detection}
\label{sec:TF}
Transfer Learning (TL) is a powerful machine learning technique that involves reassigning learned classifiers to new tasks. In other words, the knowledge acquired by the model after training on the data of the first task, encoded in the weights of the model, can be exploited by a new target task, as illustrated in Figure \ref{fig:tf}. Thus, in this work, we investigate the potential of directly using CNN models pre-trained on ImageNet dataset for fake satellite image detection. To reach this goal, for each considered model, the last FC layer is removed and the rest of the network is considered as feature extractor. Additionally, for classification purposes, three FC layers of size 1024, 512 and 2, respectively, are stacked. For the first three epochs, the weights of feature extraction, i.e., those of pre-trained network, are frizzed, after which they are fine-tuned by continuing end-to-end back propagation. This way, we retrain all layers of the model with fake satellite image dataset.

\begin{figure}[t!]
    \centering
    \includegraphics[width=0.9\linewidth]{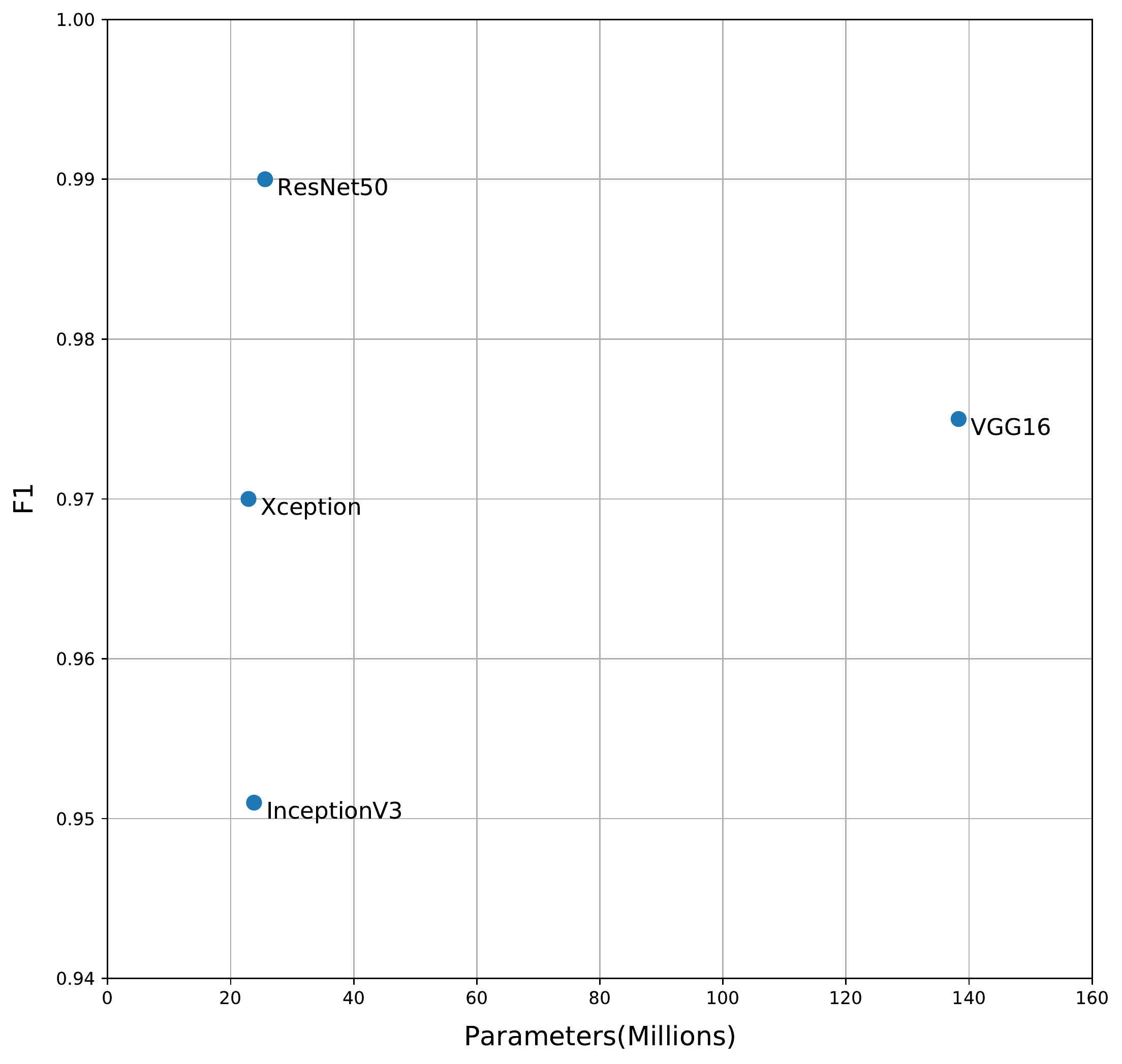}
    \caption{Trade-off between number of parameters and model's F1 score.}
    \label{fig:nb}
\end{figure}
\section{Experimental Results}
\label{sec:Exp}
To assess the performance of the four considered pre-trained CNN models, namely  VGG16 \cite{simonyan2014very}, InceptionV3 \cite{szegedy2016rethinking}, ResNet50 \cite{he2016deep} and Xception  \cite{chollet2017xception}, we experimented with the UW fake satellite image dataset \cite{zhao2021deep}. We compared the performance of these four models to seven methods combining spatial, histogram and frequency features, as proposed by Zhao \textit{et al.} \cite{zhao2021deep}. In addition, in order to assess the robustness against various image distortions, we also considered the effects of JPEG compression and Gaussian noise on the performance of each model.

The training of the models is performed with back propagation using ADAM optimization with a learning rate $lr$ of 0.0001. A dropout is applied with a rate $r$ equals to 0.2 and each mini-batch contains 32 images.

We adopted three commonly used performance criteria to evaluate the models: F1 score, precision and recall \cite{sokolova2006beyond}.
\subsection{Dataset}
\label{subsec:Data}
The UW dataset was introduced in 2021 by Zhao \textit{et al.} \cite{zhao2021deep} as the first publicly available fake satellite image dataset. The fake satellite images were generated by CycleGAN from three cities (i.e., Tacoma, Seattle and Beijing), while its authentic satellite images are collected from Google Earth’s satellite images.  UW contains 8064 satellite images, of which 4032 images are authentic color satellite images of spatial resolution $256\times256$, and their fake analog-generated images. Figure \ref{fig:uw} illustrates some fake satellite images from UW dataset.

We follow the same experimental setting as described in~\cite{zhao2021deep}. The dataset is randomly split into training (90\%) and test (10\%) sets. In order to fine-tune hyperparameters, the training set is further divided into training
(70\%) and validation (20\%) sets.
\subsection{Results and Discussion}
We compared the detection performance of the four models under 
three scenarios: 1) original images from UW dataset, 2) images distorted using JPEG compression, and finally, 3) images distorted using Gaussian noise. For the second and third scenarios, it is important to note that distortions were only applied to the test set.
\begin{table}[t!]
\caption{Detection performance of the four considered CNNs models on UW dataset (original images). The top result is highlighted in boldface.}
\label{tab:perf}
\centering
\begin{adjustbox}{max width=0.5\textwidth}
\begin{tabular}{c|c|ccc}\hline\hline
Method&Features$\backslash$Architectures  & F1 & Precision &  Recall \\ \hline
\STAB{\multirow{7}{*}{\rotatebox[origin=c]{90}{Zhao \textit{et al.} \cite{zhao2021deep}}}} &
Spatial  & 0.858 & 0.886  & 0.846  \\
&Histogram & 0.759 & 0.749  & 0.816  \\
&Frequency & 0.752 & 0.557  & 0.967 \\
&Spatial + Histogram & 0.855 & 0.861  & 0.943 \\
&Spatial + Frequency & 0.790 & 0.722  & 0.746  \\
&Histogram + Frequency  & 0.859 & 0.749  & 0.917 \\
&Spatial + Histogram + Frequency  & 0.850 & 0.848 & 0.928 \\\hdashline 
\STAB{\multirow{4}{*}{\rotatebox[origin=c]{90}{CNNs TL}}}
&VGG16        & 0.975   & 0.973    & 0.988\\
&Resnet50     & \textbf{0.990}   & \textbf{0.992}    & \textbf{0.989} \\ 
&InceptionV3 & 0.951   & 0.974    & 0.942 \\ 
&Xception     & 0.970   & 0.971    & 0.968 \\ \hline\hline
\end{tabular}
\end{adjustbox}
\end{table}
\begin{table*}[t!]
\caption{Evaluation of detection performance for images corrupted by JPEG  compression standard using three quality factors $QF$. The top result is highlighted in boldface.}
\label{tab:jpeg}
\centering
\begin{tabular}{c|c|ccc|ccc|ccc}\hline\hline

\multirow{3}{*}{Method} & \multirow{3}{*}{Features$\backslash$Architectures} &\multicolumn{9}{c}{JPEG Quality Factor}\\ \cline{3-11}
                        &  &\multicolumn{3}{c|}{$QF= 90$} & \multicolumn{3}{c|}{$QF=80$} & \multicolumn{3}{c}{$QF=70$}\\ \cline{3-11}
                        & & F1 & Precision &  Recall & F1 & Precision &  Recall& F1 & Precision &  Recall\\ \hline
Zhao \textit{et al.} \cite{zhao2021deep} & Spatial + Histogram + Frequency  & 0.873 & 0.802  & 0.956 & 0.883 & 0.802  & 0.983& 0.907 & 0.850  & 0.971 \\\hline
\multirow{4}{*}{CNNs TL}  & VGG16 & 0.979 & 0.9807  & 0.978& 0.974 & 0.980  & 0.968& 0.978 & 0.980  & 0.976 \\
                      & Resnet50 & \textbf{0.991} & 0.985  & 0.997 & \textbf{0.988} & 0.985  & 0.990 & \textbf{0.990} & 0.985  & 0.995 \\
                      & InceptionV3 & 0.934 & 0.945  & 0.923 & 0.920 & 0.944  & 0.896& 0.920 & 0.944  & 0.896 \\
                      & Xception & 0.948 & 0.972  & 0.925 & 0.933 & 0.971  & 0.899& 0.912 & 0.970  & 0.860 \\
\hline\hline
\end{tabular}
\end{table*}
\begin{table*}[htbp]
\caption{Evaluation of detection performance for images corrupted by additive white Gaussian noise using three standard deviation $\sigma$ values. The top result is highlighted in boldface.}\label{tab:noise}
\centering
\begin{tabular}{c|c|ccc|ccc|ccc}\hline\hline

\multirow{3}{*}{Method} & \multirow{3}{*}{Features$\backslash$Architectures} &\multicolumn{9}{c}{Gaussian Noise}\\ \cline{3-11}
                        &  &\multicolumn{3}{c|}{$\sigma= 0.02$} & \multicolumn{3}{c|}{$\sigma=0.06$} & \multicolumn{3}{c}{$\sigma=1$}\\ \cline{3-11}
                        & & F1 & Precision &  Recall & F1 & Precision &  Recall& F1 & Precision &  Recall\\ \hline
Zhao \textit{et al.} \cite{zhao2021deep} & Spatial + Histogram + Frequency  & 0.522 & 0.650  & 0.436 & 0.469 & 0.582  & 0.335& 0.407 & 0.578  & 0.304 \\\hline
\multirow{4}{*}{CNNs TL}  & VGG16 & \textbf{0.911} & 0.925  & 0.896 & \textbf{0.899} & 0.915  & 0.865& \textbf{0.898} & 0.914  & 0.856 \\
                      & Resnet50 & 0.798 & 0.798  & 0.788 & 0.782 & 0.775  & 0.788& 0.781 & 0.786  & 0.776 \\
                      & InceptionV3 & 0.812 & 0.695  & 0.976 & 0.803 & 0.652  & 0.976& 0.795 & 0.628  & 0.938 \\
                      & Xception & 0.764 & 0.650  & 0.928 & 0.761 & 0.635  & 0.908& 0.756 & 0.612  & 0.902 \\
\hline\hline
\end{tabular}
\end{table*}
\subsubsection{Detection performance on the original images}
Table~\ref{tab:perf} shows the detection performance on the original UW test set. From this table, we can see that the CNN-based methods achieves much better performance than the methods of Zhao \textit{et al.} \cite{zhao2021deep}. In particular, ResNet-50 achieves the best results with an F1 score of 0.990 and outperforms both InceptionV3 and Xception, although they have relatively the same number of parameters (see Figure\ref{fig:nb}). Similar high performance is achieved  using the VGG16 and Xception models. However, by comparing the results obtained by the methods based on handcrafted features, i.e., spatial, histogram, and frequency features, we can observe a significant drop in performance compared to deep learning based methods. For example, the combination of Histogram + Frequency, which is the best performing method based on handcrafted features, achieves an F1 score of 0.859.

\subsubsection{Detection performance on the JPEG compression images}
To examine the effect of compression distortion on the detection performance of benchmarking methods, all images of test set were  compressed by JPEG with three different quality factors $QF = 90$, $80$ and $70$. The results of the experiments are depicted in Table~\ref{tab:jpeg}. As can be seen from these results, the impact of JPEG compression on detection performance differs from one method to another. From Table~\ref{tab:jpeg}, it is clear that the detection performance of InceptionV3 and Xception deteriorates as the value of $QF$ decreases, while it remains nearly unchanged for both VGG16 and ResNet-50 regardless of the $QF$. Surprisingly, the detection performance of the hand-crafted-based methods improves as the value of the $QF$ decreases. 
This may be explained by the image quality metric features included in Zhao \textit{et al.} \cite{zhao2021deep} method to capture more discriminant features in the case of compressed images. Another reason is the fact that the method of Zhao \textit{et al.} considers different frequency features allowing high performance in this scenario.
\subsubsection{Detection performance on the Gaussian noise images}
Here, we measure the impact of Gaussian noise on detection performance. For this, the images of the original test set are corrupted with three $sigma$ Gaussian noise levels. The result is shown in Table~\ref{tab:noise}. As can be seen, additive Gaussian noise has a negative impact and the detection performance of all methods drop significantly on distorted images compared to the original versions. The reason is that noise is usually dominant at high frequencies, and we know that the most salient features for detecting fake images are at high frequencies. Thus, the noise disturbs this prominent information.
\section{Conclusion}
\label{sec:conclusion}
 In this work, we evaluated the efficiency of CNN-based models for the detection of fake satellite images. To this end, four CNN-based models, namely VGG16, ResNet-50, InceptionV3 and Xception, were pre-trained on ImageNet dataset and fine-tuned on UW dataset to distinguish between authentic and counterfeit satellite images.

Evaluation results in terms of F1 scores, precision, and recall revealed that the ResNet-50 outperformed all other methods, particularly InceptionV3 and Xception. Moreover, we compared the performance of the CNN-based methods against hand-crafted-based methods. Our experiments showed that CNN-based approaches significantly outperform the hand-crafted-based methods. We also analyzed the effect of image compression and noise on the detection performance. Typically, these distortions cause a significant drop in performance compared to using the original images. In particular, Gaussian noise has the greatest negative impact on the detection performance compared to JPEG compression. Surprisingly, the detection performance of the hand-crafted-based methods seems to be the least affected by image compression.
We plan to release all the data and codes for public use to support the principle of reproducible research and fair comparison.


\bibliographystyle{IEEEtran}
\bibliography{ref}

\begin{thebibliography}{10}
\providecommand{\url}[1]{#1}
\csname url@samestyle\endcsname
\providecommand{\newblock}{\relax}
\providecommand{\bibinfo}[2]{#2}
\providecommand{\BIBentrySTDinterwordspacing}{\spaceskip=0pt\relax}
\providecommand{\BIBentryALTinterwordstretchfactor}{4}
\providecommand{\BIBentryALTinterwordspacing}{\spaceskip=\fontdimen2\font plus
\BIBentryALTinterwordstretchfactor\fontdimen3\font minus
  \fontdimen4\font\relax}
\providecommand{\BIBforeignlanguage}[2]{{%
\expandafter\ifx\csname l@#1\endcsname\relax
\typeout{** WARNING: IEEEtran.bst: No hyphenation pattern has been}%
\typeout{** loaded for the language `#1'. Using the pattern for}%
\typeout{** the default language instead.}%
\else
\language=\csname l@#1\endcsname
\fi
#2}}
\providecommand{\BIBdecl}{\relax}
\BIBdecl

\bibitem{chai2021deep}
J.~Chai, H.~Zeng, A.~Li, and E.~W. Ngai, ``Deep learning in computer vision: A
  critical review of emerging techniques and application scenarios,''
  \emph{Machine Learning with Applications}, vol.~6, p. 100134, 2021.

\bibitem{goodfellow2014generative}
I.~Goodfellow, J.~Pouget-Abadie, M.~Mirza, B.~Xu, D.~Warde-Farley, S.~Ozair,
  A.~Courville, and Y.~Bengio, ``Generative adversarial nets,'' \emph{Advances
  in neural information processing systems}, vol.~27, pp. 2672--2680, 2014.

\bibitem{vincent2008extracting}
P.~Vincent, H.~Larochelle, Y.~Bengio, and P.-A. Manzagol, ``Extracting and
  composing robust features with denoising autoencoders,'' in \emph{Proceedings
  of the 25th international conference on Machine learning}, 2008, pp.
  1096--1103.

\bibitem{korshunov2018deepfakes}
P.~Korshunov and S.~Marcel, ``Deepfakes: a new threat to face recognition?
  assessment and detection,'' \emph{arXiv preprint arXiv:1812.08685}, 2018.

\bibitem{rossler2019faceforensics++}
A.~Rossler, D.~Cozzolino, L.~Verdoliva, C.~Riess, J.~Thies, and M.~Nie{\ss}ner,
  ``Faceforensics++: Learning to detect manipulated facial images,'' in
  \emph{Proceedings of the IEEE International Conference on Computer Vision},
  2019, pp. 1--11.

\bibitem{dolhansky2020deepfake}
B.~Dolhansky, J.~Bitton, B.~Pflaum, J.~Lu, R.~Howes, M.~Wang, and C.~C. Ferrer,
  ``The deepfake detection challenge (dfdc) dataset,'' \emph{arXiv preprint
  arXiv:2006.07397}, 2020.

\bibitem{ajder2019state}
H.~Ajder, G.~Patrini, F.~Cavalli, and L.~Cullen, ``The state of deepfakes:
  Landscape, threats, and impact,'' \emph{Amsterdam: Deeptrace}, 2019.

\bibitem{korshunov2021subjective}
P.~Korshunov and S.~Marcel, ``Subjective and objective evaluation of deepfake
  videos,'' in \emph{ICASSP 2021-2021 IEEE International Conference on
  Acoustics, Speech and Signal Processing (ICASSP)}.\hskip 1em plus 0.5em minus
  0.4em\relax IEEE, 2021, pp. 2510--2514.

\bibitem{marr2019best}
B.~Marr, ``The best (and scariest) examples of ai-enabled deepfakes,''
  \emph{Re-trieved from https://www. forbes.
  com/sites/bernardmarr/2019/07/22/the-best-and-scariest-examples-of-aienabled-deepfakes},
  2019.

\bibitem{chesney2019deep}
B.~Chesney and D.~Citron, ``Deep fakes: a looming challenge for privacy,
  democracy, and national security,'' \emph{Calif. L. Rev.}, vol. 107, p. 1753,
  2019.

\bibitem{vaccari2020deepfakes}
C.~Vaccari and A.~Chadwick, ``Deepfakes and disinformation: Exploring the
  impact of synthetic political video on deception, uncertainty, and trust in
  news,'' \emph{Social Media+ Society}, vol.~6, no.~1, p. 2056305120903408,
  2020.

\bibitem{tolosana2020deepfakes}
R.~Tolosana, R.~Vera-Rodriguez, J.~Fierrez, A.~Morales, and J.~Ortega-Garcia,
  ``Deepfakes and beyond: A survey of face manipulation and fake detection,''
  \emph{arXiv preprint arXiv:2001.00179}, 2020.

\bibitem{mirsky2021creation}
Y.~Mirsky and W.~Lee, ``The creation and detection of deepfakes: A survey,''
  \emph{ACM Computing Surveys (CSUR)}, vol.~54, no.~1, pp. 1--41, 2021.

\bibitem{lyu2022deepfake}
S.~Lyu, ``Deepfake detection,'' in \emph{Multimedia Forensics}.\hskip 1em plus
  0.5em minus 0.4em\relax Springer, Singapore, 2022, pp. 313--331.

\bibitem{zhao2021deep}
B.~Zhao, S.~Zhang, C.~Xu, Y.~Sun, and C.~Deng, ``Deep fake geography? when
  geospatial data encounter artificial intelligence,'' \emph{Cartography and
  Geographic Information Science}, vol.~48, no.~4, pp. 338--352, 2021.

\bibitem{tucker2019newest}
P.~Tucker, ``The newest ai-enabled weapon: Deep-faking photos of the earth,''
  \emph{Defense One.
  https://www.defenseone.com/technology/2019/03/next-phase-ai-deep-faking-whole-world-and-china-ahead/155944/},
  2019.

\bibitem{xu2018satellite}
C.~Xu and B.~Zhao, ``Satellite image spoofing: Creating remote sensing dataset
  with generative adversarial networks (short paper),'' in \emph{10th
  International conference on geographic information science (GIScience
  2018)}.\hskip 1em plus 0.5em minus 0.4em\relax Schloss
  Dagstuhl-Leibniz-Zentrum fuer Informatik, 2018.

\bibitem{zhongming2021growing}
Z.~Zhongming, L.~Linong, Y.~Xiaona, Z.~Wangqiang, L.~Wei \emph{et~al.}, ``A
  growing problem of'deepfake geography': How ai falsifies satellite images,''
  2021.

\bibitem{chen2021geo}
H.-S. Chen, K.~Zhang, S.~Hu, S.~You, and C.-C.~J. Kuo, ``Geo-defakehop:
  High-performance geographic fake image detection,'' \emph{arXiv preprint
  arXiv:2110.09795}, 2021.

\bibitem{verdoliva2020media}
L.~Verdoliva, ``Media forensics and deepfakes: an overview,'' \emph{IEEE
  Journal of Selected Topics in Signal Processing}, vol.~14, no.~5, pp.
  910--932, 2020.

\bibitem{masood2022deepfakes}
M.~Masood, M.~Nawaz, K.~M. Malik, A.~Javed, A.~Irtaza, and H.~Malik,
  ``Deepfakes generation and detection: State-of-the-art, open challenges,
  countermeasures, and way forward,'' \emph{Applied Intelligence}, pp. 1--53,
  2022.

\bibitem{agarwal2017swapped}
A.~Agarwal, R.~Singh, M.~Vatsa, and A.~Noore, ``Swapped! digital face
  presentation attack detection via weighted local magnitude pattern,'' in
  \emph{2017 IEEE International Joint Conference on Biometrics (IJCB)}.\hskip
  1em plus 0.5em minus 0.4em\relax IEEE, 2017, pp. 659--665.

\bibitem{akhtar2019comparative}
Z.~Akhtar and D.~Dasgupta, ``A comparative evaluation of local feature
  descriptors for deepfakes detection,'' in \emph{2019 IEEE International
  Symposium on Technologies for Homeland Security (HST)}.\hskip 1em plus 0.5em
  minus 0.4em\relax IEEE, 2019, pp. 1--5.

\bibitem{matern2019exploiting}
F.~Matern, C.~Riess, and M.~Stamminger, ``Exploiting visual artifacts to expose
  deepfakes and face manipulations,'' in \emph{2019 IEEE Winter Applications of
  Computer Vision Workshops (WACVW)}.\hskip 1em plus 0.5em minus 0.4em\relax
  IEEE, 2019, pp. 83--92.

\bibitem{karras2017progressive}
T.~Karras, T.~Aila, S.~Laine, and J.-k. Lehtinen, ``Progressive growing of gans
  for improved quality, stability, and variation,'' \emph{arXiv preprint
  arXiv:1710.10196}, 2017.

\bibitem{miyato2018spectral}
T.~Miyato, T.~Kataoka, M.~Koyama, and Y.~Yoshida, ``Spectral normalization for
  generative adversarial networks,'' \emph{arXiv preprint arXiv:1802.05957},
  2018.

\bibitem{bellemare2017cramer}
M.~G. Bellemare, I.~Danihelka, W.~Dabney, S.~Mohamed, B.~Lakshminarayanan,
  S.~Hoyer, and R.~Munos, ``The cramer distance as a solution to biased
  wasserstein gradients,'' \emph{arXiv preprint arXiv:1705.10743}, 2017.

\bibitem{binkowski2018demystifying}
M.~Bi{\'n}kowski, D.~J. Sutherland, M.-c. Arbel, and A.~Gretton, ``Demystifying
  mmd gans,'' \emph{arXiv preprint arXiv:1801.01401}, 2018.

\bibitem{rossler2018faceforensics}
A.~R{\"o}ssler, D.~Cozzolino, L.~Verdoliva, C.~Riess, J.~Thies, and
  M.~Nie{\ss}ner, ``Faceforensics: A large-scale video dataset for forgery
  detection in human faces,'' \emph{arXiv preprint arXiv:1803.09179}, 2018.

\bibitem{mccloskey2018detecting}
S.~McCloskey and M.~Albright, ``Detecting gan-generated imagery using color
  cues,'' \emph{arXiv preprint arXiv:1812.08247}, 2018.

\bibitem{guan2019mfc}
H.~Guan, M.~Kozak, E.~Robertson, Y.~Lee, A.~N. Yates, A.~Delgado, D.~Zhou,
  T.~Kheyrkhah, J.~Smith, and J.~Fiscus, ``Mfc datasets: Large-scale benchmark
  datasets for media forensic challenge evaluation,'' in \emph{2019 IEEE Winter
  Applications of Computer Vision Workshops (WACVW)}.\hskip 1em plus 0.5em
  minus 0.4em\relax IEEE, 2019, pp. 63--72.

\bibitem{zhang2017automated}
Y.~Zhang, L.~Zheng, and V.~L. Thing, ``Automated face swapping and its
  detection,'' in \emph{2017 IEEE 2nd International Conference on Signal and
  Image Processing (ICSIP)}.\hskip 1em plus 0.5em minus 0.4em\relax IEEE, 2017,
  pp. 15--19.

\bibitem{bayar2016deep}
B.~Bayar and M.~C. Stamm, ``A deep learning approach to universal image
  manipulation detection using a new convolutional layer,'' in
  \emph{Proceedings of the 4th ACM workshop on information hiding and
  multimedia security}, 2016, pp. 5--10.

\bibitem{rahmouni2017distinguishing}
N.~Rahmouni, V.~Nozick, J.~Yamagishi, and I.~Echizen, ``Distinguishing computer
  graphics from natural images using convolution neural networks,'' in
  \emph{2017 IEEE Workshop on Information Forensics and Security (WIFS)}.\hskip
  1em plus 0.5em minus 0.4em\relax IEEE, 2017, pp. 1--6.

\bibitem{marra2018detection}
F.~Marra, D.~Gragnaniello, D.~Cozzolino, and L.~Verdoliva, ``Detection of
  gan-generated fake images over social networks,'' in \emph{2018 IEEE
  Conference on Multimedia Information Processing and Retrieval (MIPR)}.\hskip
  1em plus 0.5em minus 0.4em\relax IEEE, 2018, pp. 384--389.

\bibitem{tariq2018detecting}
S.~Tariq, S.~Lee, H.~Kim, Y.~Shin, and S.~S. Woo, ``Detecting both machine and
  human created fake face images in the wild,'' in \emph{Proceedings of the 2nd
  international workshop on multimedia privacy and security}, 2018, pp. 81--87.

\bibitem{simonyan2014very}
K.~Simonyan and A.~Zisserman, ``Very deep convolutional networks for
  large-scale image recognition,'' \emph{arXiv preprint arXiv:1409.1556}, 2014.

\bibitem{he2016deep}
K.~He, X.~Zhang, S.~Ren, and J.~Sun, ``Deep residual learning for image
  recognition,'' in \emph{Proceedings of the IEEE conference on computer vision
  and pat-tern recognition}, 2016, pp. 770--778.

\bibitem{nguyen2019capsule}
H.~H. Nguyen, J.~Yamagishi, and I.~Echizen, ``Capsule-forensics: Using capsule
  networks to detect forged images and videos,'' in \emph{ICASSP 2019-2019 IEEE
  International Conference on Acoustics, Speech and Signal Processing
  (ICASSP)}.\hskip 1em plus 0.5em minus 0.4em\relax IEEE, 2019, pp. 2307--2311.

\bibitem{marra2019incremental}
F.~Marra, C.~Saltori, G.~Boato, and L.~Verdoliva, ``Incremental learning for
  the detection and classification of gan-generated images,'' in \emph{2019
  IEEE International Workshop on Information Forensics and Security
  (WIFS)}.\hskip 1em plus 0.5em minus 0.4em\relax IEEE, 2019, pp. 1--6.

\bibitem{szegedy2016rethinking}
C.~Szegedy, V.~Vanhoucke, S.~Ioffe, J.~Shlens, and Z.~Wojna, ``Rethinking the
  inception architecture for computer vision,'' in \emph{Proceedings of the
  IEEE conference on computer vision and pattern recognition}, 2016, pp.
  2818--2826.

\bibitem{chollet2017xception}
F.~Chollet, ``Xception: Deep learning with depthwise separable convolutions,''
  in \emph{Proceedings of the IEEE conference on computer vision and pattern
  recognition}, 2017, pp. 1251--1258.

\bibitem{russakovsky2015imagenet}
O.~Russakovsky, J.~Deng, H.~Su, J.~Krause, S.~Satheesh, S.~Ma, Z.~Huang,
  A.~Karpathy, A.~Khosla, M.~Bernstein \emph{et~al.}, ``Imagenet large scale
  visual recognition challenge,'' \emph{International journal of computer
  vision}, vol. 115, no.~3, pp. 211--252, 2015.

\bibitem{krizhevsky2012imagenet}
A.~Krizhevsky, I.~Sutskever, and G.~E. Hinton, ``Imagenet classification with
  deep convolutional neural networks,'' \emph{Advances in neural information
  processing systems}, vol.~25, 2012.

\bibitem{szegedy2015going}
C.~Szegedy, W.~Liu, Y.~Jia, P.~Sermanet, S.~Reed, D.~Anguelov, D.~Erhan,
  V.~Vanhoucke, and A.~Rabinovich, ``Going deeper with convolutions,'' in
  \emph{Proceedings of the IEEE conference on computer vision and pattern
  recognition}, 2015, pp. 1--9.

\bibitem{sokolova2006beyond}
M.~Sokolova, N.~Japkowicz, and S.~Szpakowicz, ``Beyond accuracy, f-score and
  roc: a family of discriminant measures for performance evaluation,'' in
  \emph{Australasian joint conference on artificial intelligence}.\hskip 1em
  plus 0.5em minus 0.4em\relax Springer, 2006, pp. 1015--1021.

\end{thebibliography}

\end{document}